\documentclass[runningheads]{llncs}
\usepackage{graphicx}
\usepackage{amsmath,amssymb} 
\usepackage{color}
\usepackage[width=122mm,left=12mm,paperwidth=146mm,height=193mm,top=12mm,paperheight=217mm]{geometry}
\usepackage{subfigure}
\usepackage{multirow}
\def\ie{{\em i.e.}}
\def\eg{{\em e.g.}}

\def\etal{{\em et al.}}

\begin{document}
\pagestyle{headings}
\mainmatter

\title{CASIA-SURF CeFA: A Benchmark for Multi-modal Cross-ethnicity Face Anti-spoofing} 

\author{
    Ajian Liu$^{\rm 1*}$,
    Zichang Tan$^{\rm 2}$\thanks{Equal Contribution},
    Xuan Li$^{\rm 3}$,
    Jun Wan$^{\rm 2}$\thanks{Corresponding Author, email:  jun.wan@ia.ac.cn},
    Sergio Escalera$^{\rm 4}$ \\
    Guodong Guo$^{\rm 5}$,
    Stan Z. Li$^{\rm 1,2}$\vspace{1.2mm}}


\institute{
    $^{\rm 1}$MUST, Macau, China;
    $^{\rm 2}$NLPR, CASIA, China;
    $^{\rm 3}$BJTU, China; \\
	$^{\rm 4}$CVC, UB, Spain;
	$^{\rm 5}$Baidu Research, China
    }

\maketitle

\begin{abstract}
  Ethnic bias has proven to negatively affect the performance of face recognition systems, and it remains an open research problem in face anti-spoofing. In order to study the ethnic bias for face anti-spoofing, we introduce the largest up to date CASIA-SURF Cross-ethnicity Face Anti-spoofing (CeFA) dataset (briefly named CeFA), covering $3$ ethnicities, $3$ modalities, $1,607$ subjects, and 2D plus 3D attack types. Four protocols are introduced to measure the affect under varied evaluation conditions, such as cross-ethnicity, unknown spoofs or both of them. To the best of our knowledge, CeFA is the first dataset including explicit ethnic labels in current published/released datasets for face anti-spoofing. Then, we propose a novel multi-modal fusion method as a strong baseline to alleviate these bias, namely, the static-dynamic fusion mechanism applied in each modality (\ie, RGB, Depth and infrared image). Later, a partially shared fusion strategy is proposed to learn complementary information from multiple modalities. Extensive experiments demonstrate that the proposed method achieves state-of-the-art results on the CASIA-SURF, OULU-NPU, SiW and the CeFA dataset.
\end{abstract}

\section{Introduction}
Face anti-spoofing~\cite{Boulkenafet2017Face,Liu2018Learning,shao2019multi} is a key element to avoid security breaches in face recognition systems. 
The presentation attack detection (PAD) technique is a vital stage prior to visual face recognition. Although ethnic bias has been verified to severely affect the performance of face recognition systems~\cite{race_bias_fr,AlviTurning,Wang_2019_ICCV}, it still remains an open research problem in face anti-spoofing. Based on the experiment in Section~\ref{bias}, the state-of-the-art (SOTA) algorithms also suffer from ethnic bias. More specifically, the value of ACER is at least $8\%$ higher in Central Asia than that of East Asia in Table~\ref{tab:racial_bias}. However, there is no available dataset with ethnic labels and associated protocol for its evaluation. Furthermore, as shown in Table~\ref{tab:datasets_list}, the existing face anti-spoofing datasets (\ie  CASIA-FASD~\cite{Zhang2012A}, Replay-Attack~\cite{Chingovska2012On}, OULU-NPU~\cite{Boulkenafet2017OULU} and SiW~\cite{Liu2018Learning}) have limited number of samples and most of them just contain the RGB modality. Although CASIA-SURF~\cite{DBLP:conf/cvpr/abs-1812-00408} is a large dataset in comparison to the existing alternatives, it still provides limited attack types (only 2D print attack) and single ethnicity.
\begin{figure}[!htb]
\centering
\begin{minipage}[t]{0.86\columnwidth}
\subfigure[] { \label{fig:sample_cefa}     \includegraphics[width=4.0in]{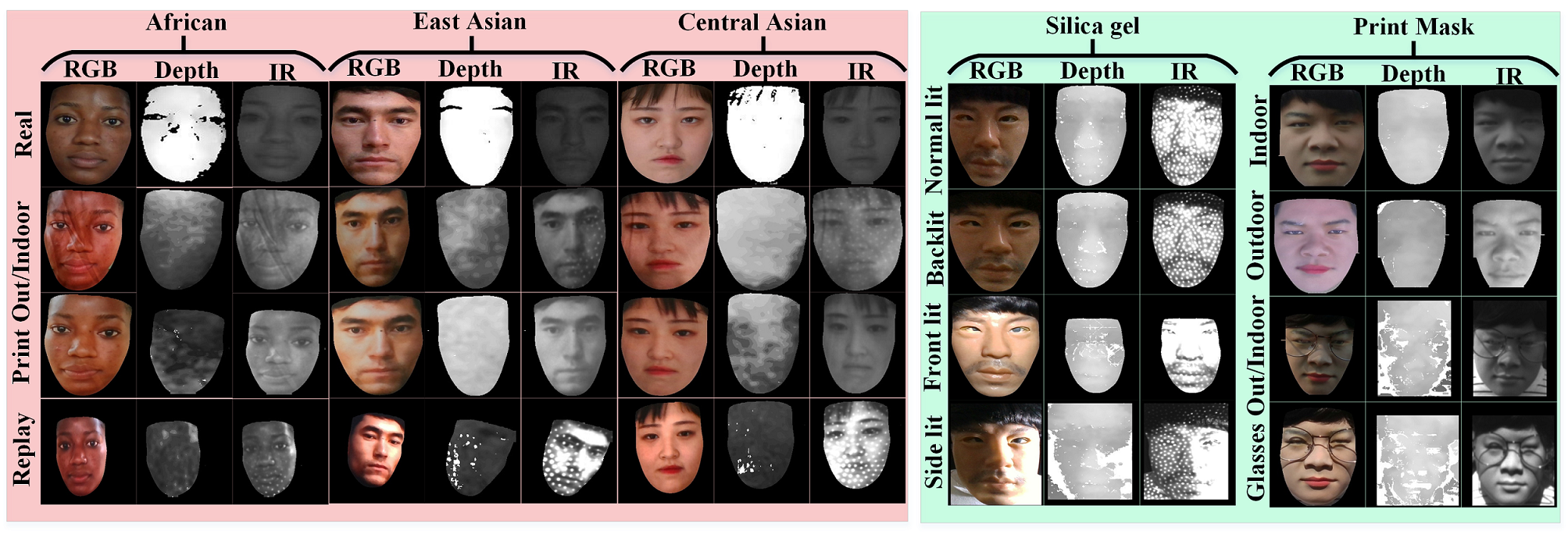} }
\end{minipage}
\begin{minipage}[t]{0.1\columnwidth}
\subfigure[] { \label{fig:gender_age}     \includegraphics[width=0.62in]{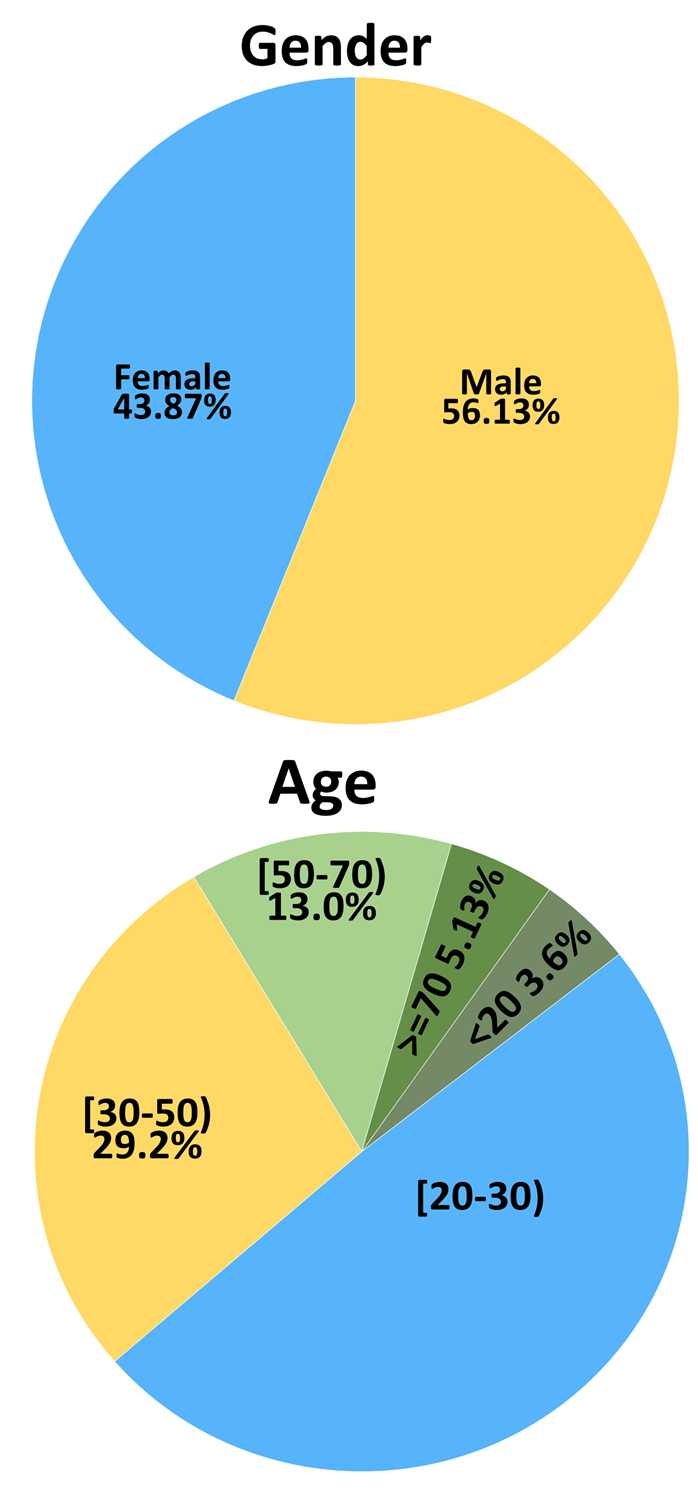} }
\end{minipage}
\caption{(a): Samples of the CeFA dataset. It contains 1,607 subjects, 3 different ethnicities (\ie, Africa, East Asia, and Central Asia) and modalities(\ie, RGB, Depth and IR), with 4 attack types (\ie, print attack, replay attack, 3D print and silica gel attacks) under various lighting conditions. Light red/blue background indicates 2D/3D attack. (b): Gender and age  distributions of the CeFA.}
\end{figure}

\newcommand{\tabincell}[2]{\begin{tabular}{@{}#1@{}}#2\end{tabular}}
\begin{table}[]
\begin{center}
\caption{Comparison of existing face PAD databases. (\textit{*} indicates the dataset only contains images. AS: Asian, A: Africa, U: Caucasian, I: Indian, E: East Asia, C: Central Asia.)}
\scalebox{0.76}{
\begin{tabular}{|c|c|c|c|c|c|c|c|}
\hline
    Dataset &Year &\#Subject &\#Num &Attack &Modality &Device &Ethnicity \\ \hline \hline
    Replay-Attack~\cite{Chingovska2012On}& 2012  & 50 & 1200 &Print,Replay & RGB & RGB Camera & -\\ \hline
    CASIA-FASD~\cite{Zhang2012A} & 2012 & 50  & 600 &Print,Cut,Replay & RGB & RGB Camera & -\\\hline
    3DMAD~\cite{ERDOGMUS_BTAS-2013} & 2014  & 17 & 255  &3D print mask& RGB/Depth& RGB Camera/Kinect  & -\\ \hline
    MSU-MFSD~\cite{Wen2015Face}  & 2015 & 35 & 440  &Print,Replay & RGB & Cellphone/Laptop & - \\ \hline
    Replay-Mobile~\cite{Costa2016The}  & 2016 & 40  & 1030  & Print,Replay & RGB & Cellphone  & -\\ \hline
    Msspoof~\cite{msspoof-2015}  & 2016  & 21 & 4704\textsuperscript{\textit{*}}& \begin{tabular}[c]{@{}c@{}}Print\end{tabular} & RGB/IR & RGB/IR Camera & -\\ \hline
    OULU-NPU~\cite{Boulkenafet2017OULU} & 2017 & 55 & 5940 &Print,Replay & RGB & \begin{tabular}[c]{@{}c@{}}RGB Camera\end{tabular} & -\\ \hline
    SiW~\cite{Liu2018Learning}  & 2018  & 165  & 4620  & \begin{tabular}[c]{@{}c@{}}Print,Replay\end{tabular} & RGB& RGB Camera  & \tabincell{c}{AS/A/\\U/I}\\ \hline
    CASIA-SURF~\cite{DBLP:conf/cvpr/abs-1812-00408}  & 2019  & 1000  & 21000  & \begin{tabular}[c]{@{}c@{}}Print,Cut\end{tabular}  & RGB/Depth/IR & Intel Realsense  & E\\ \hline
    \multirow{4}{*}{\tabincell{c}{CeFA\\(Ours)}} & \multirow{4}{*}{2019} & 1500 & 18000& Print, Replay & \multirow{3}{*}{RGB/Depth/IR} &\multirow{3}{*}{Intel Realsense} & \multirow{3}{*}{A/E/C} \\ \cline{3-5}
    &  & 99 & 5346 &  3D print mask &  & & \\ \cline{3-5}
    &  & 8 & 192 & 3D silica gel mask &  & & \\ \cline{3-8}
    &  & \multicolumn{6}{c|}{Total: \textbf{1607} subjects, \textbf{23538} videos} \\ \hline
    \end{tabular}
}
\label{tab:datasets_list}
\end{center}
\end{table}

In order to alleviate above mentioned problems, in this paper we release a Cross-ethnicity Face Anti-spoofing  dataset (CeFA), which is the largest face anti-spoofing dataset up to date in terms of ethnicities, modalities, number of subjects and attack types. The comparison with current available datasets is shown in Table~\ref{tab:datasets_list}. Concretely, attack types of the CeFA dataset are diverse, including printing from cloth, video replay attack, 3D print and silica gel attacks. More importantly, it is the first public dataset designed for exploring the impact of cross-ethnicity. Some samples are shown in Fig.~\ref{fig:sample_cefa}.

Moreover, to improve the generalization performance of unknown attack types, multi-modal PAD methods have received special attention by an increasing number of works during last two years. Some fusion methods~\cite{DBLP:conf/cvpr/abs-1812-00408,parkin2019recognizing} restrict the interactions among different modalities since they are independent before the fusion point. Therefore, it is difficult to effectively utilize the modality relatedness from the beginning of the network to its end. In this paper, we propose a Partially Shared Multi-modal Network (PSMM-Net) as a strong baseline to alleviate ethnic and attack pattern bias. On the one hand, it allows information exchange and interaction among different modalities. On the other hand, for a single-modal branch (\eg, RGB, Depth or IR), a Static and Dynamic-based Network (SD-Net) is formulated by taking the static and dynamic images as inputs, where the dynamic image is generated by rank pooling~\cite{fernando2017rank}. To sum up, the contributions of this paper are summarized as follows: (1) We release the largest face anti-spoofing dataset CeFA up to date, which includes $3$ ethnicities, $1607$ subjects and $4$ diverse 2D/3D attack types. (2) We provide a benchmark with four comprehensive evaluation protocols to measure ethnic and attack pattern bias. (3) We propose the PSMM-Net as a strong baseline to learn the fused information from single-modal and multi-modal branches. (4) Extensive experiments demonstrate that the proposed method achieves state-of-the-art results on CeFA and other $3$ public datasets. 

\section{Related work}
\subsection{Datasets}
Face recognition systems are still dealing with ethnicity bias problems~\cite{FurlFace,KlareFace,Phillips2011An,Wang_2019_ICCV}. As an effort in the direction of mitigating ethnicity bias in face recognition, Wang~\etal~\cite{Wang_2019_ICCV} have recently released a face recognition dataset containing $4$ ethnicities to be used for algorithm design. However, there is no publicly available face anti-spoofing dataset with ethnic labels. Table~\ref{tab:datasets_list} lists main features of existing face anti-spoofing datasets: (1) The maximum number of available subjects was $165$ on the SiW dataset~\cite{Liu2018Learning} before $2019$; (2) Most of the datasets just contain RGB data, such as Replay-Attack~\cite{Chingovska2012On}, CASIA-FASD~\cite{Zhang2012A}, SiW~\cite{Liu2018Learning} and OULU-NPU~\cite{Boulkenafet2017OULU}; (3) Most datasets do not provide ethnicity information, except SiW and CASIA-SURF. Although SiW provides four ethnicities, it has neither a clear ethnic label nor a standard protocol for measuring ethnic bias in algorithms. This limitation also holds for the CASIA-SURF dataset.

\subsection{Methods}
{\flushleft \textbf{Static and Temporal Methods.}} In addition to some works~\cite{Pan2011Monocular,Komulainen2014Context,Boulkenafet2016Face} based on static texture feature learning, some temporal-based methods~\cite{Pan2007Eyeblink,patel2016cross,KollreiderReal} also have been proposed, which require from a constrained human interaction. However, these methods become vulnerable if someone presents a replay attack.
There are also methods~\cite{Boulkenafet2017Face,komulainen2013complementary} relying on more general temporal features by simply concatenating the features of consecutive frames~\cite{Wei2009A,Agarwal2016Face}. However, these algorithms are not accurate enough because of the use of hand-crafted features, such as HOG~\cite{yang2013face}, LBP~\cite{maatta2011face,de2013can}, SIFT~\cite{patel2016secure} or SURF~\cite{Boulkenafet2016Face}. Liu~\etal~\cite{Liu2018Learning} proposed a CNN-RNN model to estimate Photoplethysmography (rPPG) signals which can be detected from real but not spoof with sequence-wise supervision. Yang~\etal~\cite{yang2019face} proposed a spatio-temporal attention mechanism to fuse global temporal and local spatial information. Although these face PAD methods achieve near-perfect performance in intra-database experiments, they are vulnerable when facing complex lighting environments in practical applications. Inspired by~\cite{feng2016integration}, we feed the static and dynamic images to SD-Net, which the dynamic image generated by rank pooling~\cite{fernando2017rank} instead of optical flow map~\cite{feng2016integration}. Additionally, our SD-Net not only captures the static and dynamic features, but also static-dynamic fusion features in an end-to-end way.

{\flushleft\textbf{Multi-modal Fusion Methods.}} Zhang~\etal~\cite{DBLP:conf/cvpr/abs-1812-00408} proposed a fusion network with $3$ streams using ResNet-18 as the backbone, where each stream is used to extract low level features from RGB, Depth and IR data, respectively. Then, these features are concatenated and passed to the last two residual blocks. Similar to~\cite{DBLP:conf/cvpr/abs-1812-00408}, Tao~\etal~\cite{shen2019facebagnet} proposed a multi-stream CNN architecture called FaceBagNet, which uses patch-level images as input and modality feature erasing (MFE) operation to prevent from overfitting. All previous methods just consider as a key fusion component the concatenation of features from multiple modalities. Unlike~\cite{DBLP:conf/cvpr/abs-1812-00408,parkin2019recognizing,shen2019facebagnet}, we propose the PSMM-Net, where three modality-specific networks and one shared network are connected by using a partially shared structure to learn discriminative fused features for face anti-spoofing.

\section{CeFA dataset}
In this section, we introduce the CeFA dataset, including acquisition details, attack types, and proposed evaluation protocols.
{\flushleft \textbf{Acquisition Details.}} We use the Intel Realsense to capture the RGB, Depth and IR videos simultaneously at $30 fps$. The resolution is $1280$ $\times$ $720$ pixels for each video frame. Subjects are asked to move smoothly their head so as to have a maximum of around $30^0$ deviation of head pose in relation to the frontal view. Data pre-processing is similar to the one performed in~\cite{DBLP:conf/cvpr/abs-1812-00408}, except that PRNet~\cite{DBLP:conf/eccv/FengWSWZ18} is replaced by 3DFFA~\cite{zhu2017face} for face region detection. Examples of processed face regions for different visual modalities are shown in Fig.~\ref{fig:sample_cefa}.

{\flushleft \textbf{Statistics.}} As shown in Table~\ref{tab:datasets_list}, CeFA consists of 2D and 3D attack subsets. As shown in Fig.~\ref{fig:sample_cefa}, For the 2D attack subset, it consists of print and video-replay attacks captured by subjects from three ethnicities (\eg, African, East Asian and Central Asian). Each ethnicity has 500 subjects. Each subject has $1$ real sample, $2$ fake samples of print attack captured in indoor and outdoor, and $1$ fake sample of video-replay. In total, there are $18000$ videos ($6000$ per modality). The age and gender statistics for the 2D attack subset of CeFA is shown in Fig.~\ref{fig:gender_age}.

For the 3D attack subset, it has 3D print mask and silica gel face attacks. Some samples are shown in Fig.~\ref{fig:sample_cefa}. In the part of 3D print mask, it has $99$ subjects, each subject with $18$ fake samples captured in three attacks and six lighting environments. 3D print includes only face mask, wearing a wig with glasses, and wearing a wig without glasses. Lighting conditions include outdoor sunshine, outdoor shade, indoor side light, indoor front light, indoor backlit and indoor regular light. In total, there are $5346$ videos ($1782$ per modality). For silica gel face attacks, it has $8$ subjects, each subject has $8$ fake samples captured in two attacks styles and four lighting environments. Attacks include wearing a wig with glasses and wearing a wig without glasses. Lighting environments include indoor side light, indoor front light, indoor backlit and indoor normal light. In total, there are $196$ videos ($64$ per modality).

{\flushleft\textbf{Evaluation Protocols.}} The motivation of CeFA dataset is to provide a benchmark to allow for the evaluation of the generalization performance of new PAD methods in three main aspects: cross-ethnicity, cross-modality and cross-attacks. We design four protocols for the 2D attacks subset, as shown in Table~\ref{tab:protocol}, totalling $11$ sub-protocols (1\_1, 1\_2, 1\_3, 2\_1, 2\_2, 3\_1, 3\_2, 3\_3, 4\_1, 4\_2, and 4\_3). We divide $500$ subjects per ethnicity into three subject-disjoint subsets (second and fourth columns in Table~\ref{tab:protocol}). Each protocol has three data subsets: training, validation and testing sets, which contain $200$, $100$, and $200$ subjects, respectively.
\begin{table}[]
\begin{center}
\caption{Four protocols are defined for CeFA: (1) cross-ethnicity, (2) cross-PAI, (3) cross-modality, (4) cross-ethnicity\&PAI. Note that the 3D attacks subset are included in each testing protocol (not shown in the table). \& indicates merging; $*\_*$ corresponds to the name of sub-protocols. R: RGB, D: Depth, I: IR. Other abbreviated same as in Table~\ref{tab:datasets_list}.}
\scalebox{0.7}{
	\begin{tabular}{|c|c|c|c|c|c|c|c|c|c|c|c|c|c|}
		\hline
		Prot. & Subset & \multicolumn{3}{c|}{Ethnicity} & Subjects & \multicolumn{3}{c|}{Modalities} & \multicolumn{2}{c|}{PAIs} & \# real videos & \# fake videos & \# all videos \\ \hline \hline
		\multicolumn{2}{|c|}{} & 1\_1 & 1\_2 & 1\_3 & \multicolumn{9}{c|}{} \\ \hline
		\multirow{3}{*}{1} & Train & A & C & E & 1-200 & \multicolumn{3}{c|}{R\&D\&I} & \multicolumn{2}{c|}{Print\&Replay} & 600/600/600 & 1800/1800/1800 & 2400/2400/2400 \\ \cline{2-14}
		& Valid & A & C & E & 201-300 & \multicolumn{3}{c|}{R\&D\&I} & \multicolumn{2}{c|}{Print\&Replay} & 300/300/300 & 900/900/900 & 1200/1200/1200 \\ \cline{2-14}
		& Test & C\&E & A\&E & A\&C & 301-500 & \multicolumn{3}{c|}{R\&D\&I} & \multicolumn{2}{c|}{Print\&Replay} & 1200/1200/1200 & 6600/6600/6600 & 7800/7800/7800 \\ \hline
		\multicolumn{9}{|c|}{} & 2\_1 & 2\_2 & \multicolumn{3}{c|}{} \\ \hline
		\multirow{3}{*}{2} & Train & \multicolumn{3}{c|}{A\&C\&E} & 1-200 & \multicolumn{3}{c|}{R\&D\&I} & Print & Replay & 1800/1800 & 3600/1800 & 5400/3600 \\ \cline{2-14}
		& Valid & \multicolumn{3}{c|}{A\&C\&E} & 201-300 & \multicolumn{3}{c|}{R\&D\&I} & Print & Replay & 900/900 & 1800/900 & 2700/1800 \\ \cline{2-14}
		& Test & \multicolumn{3}{c|}{A\&C\&E} & 301-500 & \multicolumn{3}{c|}{R\&D\&I} & Replay & Print & 1800/1800 & 4800/6600 & 6600/8400 \\ \hline
		\multicolumn{6}{|c|}{} & 3\_1 & 3\_2 & 3\_3 & \multicolumn{5}{c|}{} \\ \hline
		\multirow{3}{*}{3} & Train & \multicolumn{3}{c|}{A\&C\&E} & 1-200 & R & D & I & \multicolumn{2}{c|}{Print\&Replay} & 600/600/600 & 1800/1800/1800 & 2400/2400/2400 \\ \cline{2-14}
		& Valid & \multicolumn{3}{c|}{A\&C\&E} & 201-300 & R & D & I & \multicolumn{2}{c|}{Print\&Replay} & 300/300/300 & 900/900/900 & 1200/1200/1200 \\ \cline{2-14}
		& Test & \multicolumn{3}{c|}{A\&C\&E} & 301-500 & D\&I & R\&I & R\&D & \multicolumn{2}{c|}{Print\&Replay} & 1200/1200/1200 & 5600/5600/5600 & 6800/6800/6800 \\ \hline
		\multicolumn{2}{|c|}{} & 4\_1 & 4\_2 & 4\_3 & \multicolumn{9}{c|}{} \\ \hline
		\multirow{3}{*}{4} & Train & A & C & E & 1-200 & R & D & I & \multicolumn{2}{c|}{Replay} & 600/600/600 & 600/600/600 & 1200/1200/1200 \\ \cline{2-14}
		& Valid & A & C & E & 201-300 & R & D & I & \multicolumn{2}{c|}{Replay} & 300/300/300 & 300/300/300 & 600/600/600 \\ \cline{2-14}
		& Test & C\&E & A\&E & A\&C & 301-500 & R & D & I & \multicolumn{2}{c|}{Print} & 1200/1200/1200 & 5400/5400/5400 & 6600/6600/6600 \\ \hline
\end{tabular}
}
\label{tab:protocol}
\end{center}
\end{table}
\\
\textbf{$\bullet$ Protocol 1 (cross-ethnicity)}: Most of the public face PAD datasets lack of ethnicity labels or do not provide with a protocol to perform cross-ethnicity evaluation. Therefore, we design the first protocol to evaluate the generalization of PAD methods for cross-ethnicity testing. One ethnicity is used for training and validation, and the left two ethnicities are used for testing. Therefore, there are three different evaluations (third column of Protocol $1$ in Table~\ref{tab:protocol}).
\\
\textbf{$\bullet$ Protocol 2 (cross-PAI)}: Given the diversity and unpredictability of attack types from different presentation attack instruments (PAI), it is necessary to evaluate the robustness of face PAD algorithms to this kind of variations (sixth column of Protocol $2$ in Table~\ref{tab:protocol}).
\\
\textbf{$\bullet$ Protocol 3 (cross-modality)}: Given the release of affordable devices capturing complementary visual modalities (\ie, Intel Resense, Mircrosoft Kinect), recently the multi-modal face anti-spoofing dataset was proposed~\cite{DBLP:conf/cvpr/abs-1812-00408}. However, there is no standard protocol to explore the generalization of face PAD methods when different train-test modalities are considered for evaluation. We define three cross-modality evaluations, each of them having one modality for training and the two remaining ones for testing (fifth column of Protocol $3$ in Table~\ref{tab:protocol}).
\\
\textbf{$\bullet$ Protocol 4 (cross-ethnicity \& PAI)}: The most challenging protocol is designed via combining the condition of both Protocol $1$ and $2$. As shown in Protocol $4$ of Table.~\ref{tab:protocol}, the testing subset introduces two unknown target variations simultaneously. Like~\cite{Boulkenafet2017OULU}, the mean and variance of evaluated metrics for these four protocols are calculated in our experiments. Detailed statistics for the different protocols are shown in Table~\ref{tab:protocol}.

\section{Proposed Method}
Here, we propose a novel strong baseline to evaluate the proposed CeFA dataset. First, the SD-Net is proposed to process the single-modal data, which is formulated by taking the static and dynamic images as
inputs. The dynamic images are generated by rank pooling. Then, the PSMM-Net is presented by learning the fusion features from multiple modalities.
\subsection{SD-Net for Single Modality}
{\flushleft \textbf{Single-modal Dynamic Image Construction.}} Rank pooling~\cite{fernando2017rank,wang2018cooperative} defines a rank function that encodes a video into a feature vector. The learning process can be seen as a convex optimization problem using the RankSVM~\cite{smola2004tutorial} formulation in Eq.\ref{Eq:ranksvm}. Let RGB (Depth or IR) video sequence with $K$ frames be represented as $<{\bf I}_1, {\bf I}_2, ..., {\bf I}_i,...,{\bf I}_K>$, and ${\bf I}_i$ denote the average of RGB (Depth or IR) features over time up to $i$-frame. The process is formulated below.
\begin{equation}\label{Eq:ranksvm}
\begin{split}
\underset{{\bf d}}{argmin}
 & \frac{1}{2} \| {\bf d} \|^2 +  \delta \times \sum_{i>j}{\xi_{ij}}  \\
 s.t. \ {\bf d}^T \cdot & ({\bf I}_i-{\bf I}_j) \geq 1- \xi_{ij}, \ \xi_{ij} \geq 0
\end{split}
\end{equation}
where $\xi_{ij}$ is the slack variable, and $\delta=\frac{2}{K(K-1)}$. By optimizing Eq.~\ref{Eq:ranksvm}, we map a sequence of $K$ frames to a single vector ${\bf d}$. In this paper, rank pooling is directly applied on the pixels of RGB (Depth or IR) frames and the dynamic image ${\bf d}$ is of the same size as the input frames. In our case, given input frame, we compute its dynamic image online with rank pooling using $K$ consecutive frames. Our selection of dynamic images for rank pooling in SD-Net is further motivated by the fact that dynamic images have proved its superiority to regular optical flow~\cite{wang2017ordered,fernando2017rank}.
\begin{figure}[t]
\begin{center}
\includegraphics[width=1.0\linewidth]{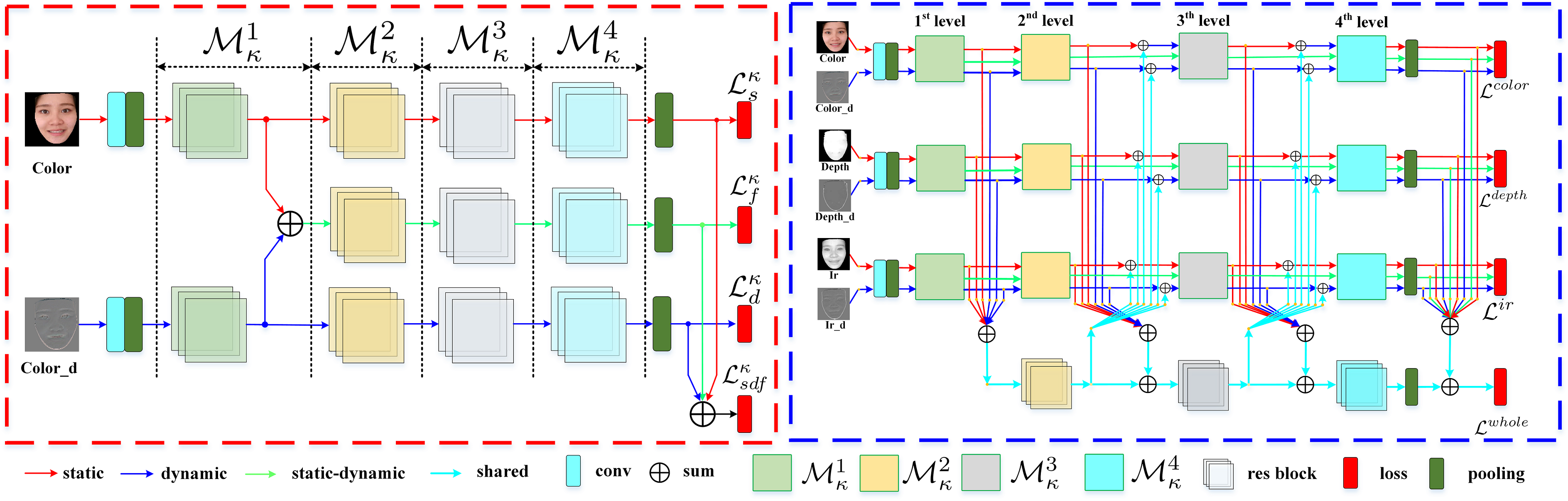}
\caption{SD-Net diagram (red box), showing a single-modal (takes RGB as an example) static-dynamic network with three branches: static (red arrow), dynamic (blue arrow) and static-dynamic (green arrow). PSMM-Net diagram (blue box) consists of two main parts: (1) Modality-specific network, which contains three SD-Nets; (2)A shared branch for all modalities, which aims to learn the complementary features among different modalities (best viewed in color).}
\label{fig:static_dynamic}
\end{center}
\end{figure}
{\flushleft \textbf{Single-modal SD-Net.}} As shown in Fig.~\ref{fig:static_dynamic}, taking the RGB modality as an example, we propose the SD-Net to learn hybrid features from static and dynamic images. It contains $3$ branches: static, dynamic and static-dynamic branches, which learn complementary features. The network takes ResNet-18~\cite{he2016deep} as the backbone. For static and dynamic branches, each of them consists of $5$ blocks (\ie, conv, res1, res2, res3, res4) and $1$ Global Average Pooling (GAP) layer, while in the static-dynamic branch, the conv and res1 blocks are removed because it takes fused features of res1 blocks from static and dynamic branches as input.

For convenience of terminology with the rest of the paper, we divide residual blocks of the network into a set of modules $ \{ {\cal M}^{t}_{\kappa} \}_{t=1}^{4}$ according to feature level, where $\kappa \in \{ color, depth, ir \}$ is an indicator of the modality and $t$ represents the feature level. Except for the first module ${\cal M}^{1}_{\kappa}$, each module extracts static, dynamic and static-dynamic features by using a residual block, denoted as ${\bf X}^{t}_{s,\kappa}$, ${\bf X}^{t}_{d,\kappa}$ and ${\bf X}^{t}_{f,\kappa}$, respectively. The output features from each module are used as the input for the next module. The static-dynamic features ${\bf X}^{1}_{f,\kappa}$ of the first module are obtained by directly summing ${\bf X}^{1}_{s,\kappa}$ and ${\bf X}^{1}_{d,\kappa}$.

In order to ensure each branch learns independent features, each branch employs an independent loss function after the GAP layer~\cite{tan-ijcai}. In addition, a loss function based on the summed features from all three branches is employed. The binary cross-entropy loss is used as the loss function. All branches are jointly and concurrently optimized to capture discriminative and complementary features for face anti-spoofing in image sequences. The overall objective function of SD-Net for the $\kappa^{th}$ modality is defined as:
\begin{equation}\label{Eq:single_modality_loss}
\begin{split}
\mathcal{L}^{\kappa}=  \mathcal{L}_{s}^{\kappa} + \mathcal{L}_{d}^{\kappa} + \mathcal{L}_{f}^{\kappa} + \mathcal{L}_{sdf}^{\kappa}
\end{split}
\end{equation}
where $\mathcal{L}_{s}^{\kappa}$, $\mathcal{L}_{d}^{\kappa}$, $\mathcal{L}_{f}^{\kappa}$ and $\mathcal{L}_{sdf}^{\kappa}$ are the losses for static branch, dynamic branch, static-dynamic branch, and summed features from all three branches of the network, respectively.

\subsection{PSMM-Net for Multi-modal Fusion}
The architecture of the proposed PSMM-Net is shown in Fig.~\ref{fig:static_dynamic}(b). It consists of two main parts: a) the modality-specific network, which contains three SD-Nets to learn features from RGB, Depth, IR modalities, respectively; b) and a shared branch for all modalities, which aims to learn the complementary features among different modalities. For the shared branch, we adopt ResNet-18, removing the first conv layer and res1 block. In order to capture correlations and complementary semantics among different modalities, information exchange and interaction among SD-Nets and the shared branch are designed. This is done in two different ways: a) forward feeding of fused SD-Net features to the shared branch, and b) backward feeding from shared branch modules output to SD-Net block inputs.

{\flushleft \textbf{Forward Feeding.}} We fuse static and dynamic SD-Nets features from all modality branches and fed them as input to its corresponding shared block. The fused process at $t^{th}$ feature level can be formulated as:
\begin{equation}\label{Eq:sum1}
  {\tilde {\bf S}}^{t} = \sum_{\kappa }{ {\bf X}}^{t}_{s,\kappa}   + \sum_{\kappa }{ {\bf X}}^{t}_{d,\kappa} + {\bf S}^t \quad t=1,2,3
\end{equation}
In the shared branch, ${\tilde {\bf S}}^{t}$ denotes the input to the $(t + 1)^{th}$ block, and ${\bf S}^t$ denotes the output of the $t^{th}$ block. Note that the first residual block is removed from the shared branch, thus ${\bf S}^1$ equals to zero.

{\flushleft \textbf{Backward Feeding.}} Shared features ${\bf S}^t$ are delivered back to the SD-Nets of the different modalities. The static features ${ {\bf X}}^{t}_{s,\kappa}$ and dynamic features ${ {\bf X}}^{t}_{d,\kappa}$
add with ${\bf S}^t$ for feature fusion. This can be denoted as:
\begin{equation}\label{Eq:sum2}
  \widetilde{\bf{X}}_{s,\kappa}^t = {\bf{X}}_{s,\kappa}^t + {{\bf{S}}^t}, \quad \widetilde{\bf{X}}_{d,\kappa}^t = {\bf{X}}_{d,\kappa}^t + {{\bf{S}}^t}
\end{equation}
where $t$ ranges from $2$ to $3$. After feature fusion, ${ \widetilde {\bf X}}^{t}_{s,\kappa}$ and ${ \widetilde {\bf X}}^{t}_{d,\kappa}$ become the new static and dynamic features, which are then feed to the next module ${\cal M}^{t+1}_{\kappa}$. Note that the exchange and interaction among SD-Nets and the shared branch are only performed for static and dynamic features. This is done to avoid hybrid features among static and dynamic information to be disturbed by multi-modal semantics.

{\flushleft \textbf{Loss Optimization.}} There are two main kind of losses employed to guide the training of PSMM-Net. The first corresponds to the losses of the three SD-Nets, \ie color, depth and ir modalities, denoted as ${\cal L}^{color}$, ${\cal L}^{depth}$ and ${\cal L}^{ir}$, respectively. The second corresponds to the loss that guides the entire network training, denoted as ${\cal L}^{whole}$, which bases on the summed features from all SD-Nets and the shared branch. The overall loss $\cal L$ of PSMM-Net is denoted as:
\begin{equation}\label{Eq:multi_modality_loss}
\begin{split}
\mathcal{L}= \mathcal{L}^{whole} + \mathcal{L}^{color} +  \mathcal{L}^{depth} + \mathcal{L}^{ir}
\end{split}
\end{equation}

\section{Experiments}
In this section, we conduct a series of experiments on CeFA and public available face anti-spoofing datasets to show the significance of the presented dataset and the effectiveness of our methodology.

\subsection{Datasets \& Metrics}
We evaluate the performance of PSMM-Net on two multi-modal (\ie, RGB, Depth and IR) datasets: CeFA and CASIA-SURF~\cite{DBLP:conf/cvpr/abs-1812-00408}, while evaluate the SD-Net on two single-modal (\ie, RGB) face anti-spoofing benchmarks: OULU-NPU~\cite{Boulkenafet2017OULU} and SiW~\cite{Liu2018Learning}.
In order to perform a consistent evaluation with prior works, we report the experimental results using the following metrics based on respective official protocols: Attack Presentation Classification Error Rate (APCER)~\cite{metrics}, Bona Fide Presentation Classification Error Rate (BPCER), Average Classification Error Rate (ACER), and Receiver Operating Characteristic (ROC) curve~\cite{DBLP:conf/cvpr/abs-1812-00408}.

\subsection{Implementation Details}
The proposed PSMM-Net is implemented with Tensorflow~\cite{AbadiTensorFlow} and run on a single NVIDIA TITAN X GPU. We resize the cropped face region to $112\times112$, and use random rotation within the range of [$-180^0$, $180^0$], flipping, cropping and color distortion for data augmentation. All models are trained for $25$ epochs via Adaptive Moment Estimation (Adam) algorithm and initial learning rate of $0.1$, which is decreased after $15$ and $20$ epochs with a factor of $10$. The batch size of each CNN stream is $64$, and the length of the consecutive frames used to construct dynamic map is set to $7$ by our experimental experience.

\subsection{Performance Biases of Diversity Ethnicities} \label{bias}
In this section, we investigate the performance biases of different ethnicities with two SOTA algorithms on the three ethnicities of CeFA. MS-SEF~\cite{DBLP:conf/cvpr/abs-1812-00408} is trained on CASIA-SURF for the multi-modal data while FAS-BAS~\cite{Liu2018Learning} is trained for the RGB data on OULU-NPU. Then, the trained models are tested on CeFA. The results are shown in Table~\ref{tab:racial_bias}. Results show that both methods behave differently for the three ethnicities, \ie, East Asian ($11.4\%$) versus Center Asian ($19.6\%$) for MS-SEF and African ($14.2\%$) versus Center Asian ($26.1\%$) for MS-SEF under the ACER metric. In addition, both methods achieve relatively good results on East Asians (\eg, the values of ACER are $11.4\%$, $15.4\%$, respectively) because of most of the samples belong to East Asians on CASIA-SURF and OULU-NPU datasets. This indicates that existing single-ethnic anti-spoofing datasets limit the ethnic generalization performance of existing methods.
\begin{table}[]
\begin{center}
\caption{Ethnic bias in deep face anti-spoofing methods. The ACER(\%) on three ethnicities are given.}
\scalebox{1.0}{
\begin{tabular}{|c|c|c|c|c|c|}
\hline
\multirow{2}{*}{Method} & \multirow{2}{*}{Trained Dataset} & \multirow{2}{*}{Modality} & \multicolumn{3}{c|}{Ethnicity(ACER\%)} \\ \cline{4-6}
&   &  & Africa   & Central Asia   & East Asia  \\ \hline
MS-SEF~\cite{DBLP:conf/cvpr/abs-1812-00408}
& CASIA-SURF~\cite{DBLP:conf/cvpr/abs-1812-00408} & RGB\&Depth\&IR & 13.9 & 19.6 & 11.4 \\ \hline
FAS-BAS~\cite{Liu2018Learning}
& OULU-NPU~\cite{Boulkenafet2017OULU} & RGB  & 14.2    & 26.1   & 15.4  \\ \hline
\end{tabular}
}
\label{tab:racial_bias}
\end{center}
\end{table}
\subsection{Baseline Model Evaluation}
 Here, we provide a benchmark for CeFA based on the proposed method. From Table~\ref{tab:baseline_protocol}, we can draw the following conclusions: (1) The ACER scores of three sub-protocols in Protocol $1$ are $0.6\%$, $4.4\%$ and $1.5\%$, respectively, which indicate the necessity to study the generalization of the face PAD methods for different ethnicities; (2) In the case of Protocol $2$, when print attack is used for training/validation and video-replay and 3D mask are used for testing, the ACER score is $0.4\%$ (sub-protocol 2\_1). When video-replay attack is used for training/validation, and print attack and 3D attack are used for testing, the ACER score is $7.5\%$ (sub-protocol 2\_2). The large gap between the results caused by the different PAI (\ie, different displays and printers). (3) Protocol $3$ evaluates cross-modality. The best result is achieved for sub-protocol 3\_1 (ACER=$4.9\%$). (4) Protocol $4$ is the most difficult evaluation scenario, which simultaneously considers cross-ethnicity and cross-PAI. All sub-protocols achieve low  performance, highlighting the challenges of our dataset: $24.5\%$, $43.2\%$, and $27.7\%$ ACER scores for 4\_1, 4\_2, and 4\_3, respectively.
\begin{table}[]
\begin{center}
\caption{PSMM-Net evaluation on the four protocols of CeFA dataset, where A$\_$B represents sub-protocol B from Protocol A, and Avg$\pm$Std indicates the mean and variance operation.}
\scalebox{0.8}{
\begin{tabular}{|c|c|c|c|c|}
\hline
\multicolumn{2}{|c|}{Protocol name}        & APCER(\%)    & BPCER(\%)   & ACER(\%)     \\ \hline \hline
\multirow{4}{*}{Protocol 1} & 1\_1     & 0.5      & 0.8     & 0.6      \\ \cline{2-5}
                            & 1\_2     & 4.8      & 4.0     & 4.4      \\ \cline{2-5}
                            & 1\_3     & 1.2      & 1.8     & 1.5      \\ \cline{2-5}
                            & Avg$\pm$Std & 2.2$\pm$2.3  & 2.2$\pm$1.6 & 2.2$\pm$2.0  \\ \hline
\multirow{3}{*}{Protocol 2} & 2\_1     & 0.1      & 0.7     & 0.4      \\ \cline{2-5}
                            & 2\_2     & 13.8     & 1.2     & 7.5      \\ \cline{2-5}
                            & Avg$\pm$Std & 7.0$\pm$9.7  & 1.0$\pm$0.4 & 4.0$\pm$5.0  \\ \hline
\multirow{4}{*}{Protocol 3} & 3\_1     & 8.9      & 0.9     & 4.9      \\ \cline{2-5}
                            & 3\_2     & 22.6     & 4.6     & 13.6     \\ \cline{2-5}
                            & 3\_3     & 21.1     & 2.3     & 11.7     \\ \cline{2-5}
                            & Avg$\pm$Std & 17.5$\pm$7.5 & 2.6$\pm$1.9 & 10.1$\pm$4.6 \\ \hline
\multirow{4}{*}{Protocol 4} & 4\_1    & 33.3      & 15.8     & 24.5      \\ \cline{2-5}
                            & 4\_2     & 78.2      & 8.3     & 43.2      \\ \cline{2-5}
                            & 4\_3     & 50.0      & 5.5     & 27.7      \\ \cline{2-5}
                            & Avg$\pm$Std & 53.8$\pm$22.7  & 9.9$\pm$5.3 & 31.8$\pm$10.0  \\ \hline
\end{tabular}
}
\label{tab:baseline_protocol}
\end{center}
\end{table}
\subsection{Ablation Analysis}
To verify the performance of our proposed baseline in alleviating ethnic bias, we perform a series of ablation experiments on Protocol $1$ (cross-ethnicity) of the CeFA dataset.

{\flushleft \textbf{Static and Dynamic Features.}} We evaluate S-Net (Static branch of SD-Net), D-Net (Dynamic branch of SD-Net) and SD-Net in this experiment. Results for RGB, Depth and IR modalities are shown in Table~\ref{tab:ablation_SDNet}. Compared to S-Net and D-Net, SD-Net achieves superior performance showing that the learned hybrid features from static and dynamic images can alleviate ethnic bias. Concretely, for RGB, Depth and IR modalities, ACER of SD-Net is $12.6\%$, $6.1\%$, $6.4\%$, versus $17.2\%$, $7.7\%$, $9.4\%$ of S-Net (improved by $4.6\%$, $1.6\%$, $3.4\%$) and $19.9\%$, $9.4\%$, $11.3\%$ of D-Net (improved by $7.3\%$, $3.3\%$, $4.9\%$), respectively. It also shows that the performance of Depth and IR modalities are superior to the RGB modality because of the variability of lighting conditions interfering with feature learning of RGB samples.
\begin{table}[]
\begin{minipage}[t]{1.0\linewidth}
\makeatletter\def\@captype{table}\makeatother\caption{Each modality group (RGB, Depth and IR) contains three experiments: static, dynamic and static-dynamic branch. Best results are shown in bold.}
\centering
\scalebox{0.74}{
\begin{tabular}{|c|c|c|c|c|c|c|c|c|c|}
    \hline
    \multirow{2}{*}{Prot.1}
        		& \multicolumn{3}{c|}{RGB}
        		& \multicolumn{3}{c|}{Depth}
        		& \multicolumn{3}{c|}{IR} \\ \cline{2-10}
        		& APCER($\%$)  & BPCER($\%$)  & ACER($\%$)  & APCER($\%$)   & BPCER($\%$)
        		& ACER($\%$)   & APCER($\%$)  & BPCER($\%$)  & ACER($\%$)  \\ \hline \hline
        		S-Net & 28.1$\pm$3.6          & \textbf{6.4$\pm$4.6}  & 17.2$\pm$3.6
        		& \textbf{5.6$\pm$3.0}  & 9.8$\pm$4.2           & 7.7$\pm$3.5
        		& 11.4$\pm$2.1          & 8.2$\pm$1.2           & 9.8$\pm$1.7  \\ \hline
        		
        		D-Net  & 20.6$\pm$4.0         & 19.3$\pm$9.0          & 19.9$\pm$4.0
        		& 11.2$\pm$5.1         & 7.5$\pm$1.5           & 9.4$\pm$2.0
        		& 8.1$\pm$1.8          & 14.4$\pm$3.8          & 11.3$\pm$2.1  \\ \hline
        		
        		SD-Net  & \textbf{14.9$\pm$6.0}         & 10.3$\pm$1.8                  & \textbf{12.6$\pm$3.4}
        		& 7.0$\pm$8.1                   & \textbf{5.2$\pm$3.5}          & \textbf{6.1$\pm$5.4}
        		& \textbf{7.3$\pm$1.2}          & \textbf{5.5$\pm$1.8}          & \textbf{6.4$\pm$1.3}  \\ \hline
\end{tabular}
}
\label{tab:ablation_SDNet}
\end{minipage}
\\
\begin{minipage}[t]{0.5\linewidth}
\makeatletter\def\@captype{table}\makeatother\caption{Effect of multiple modalities.}
\scalebox{0.76}{
\begin{tabular}{|c|c|c|c|}
\hline
        	\multirow{2}{*}{Prot.1} & \multicolumn{3}{c|}{PSMM-Net} \\ \cline{2-4}
        	& APCER($\%$) & BPCER($\%$) & ACER($\%$) \\ \hline \hline
        	RGB& 14.9$\pm$6.0 &10.3$\pm$1.8  & 12.6$\pm$3.4 \\ \hline
        	RGB\&Depth &2.3$\pm$2.9  &9.2$\pm$5.9  & 5.7$\pm$3.5 \\ \hline
        	RGB\&Depth\&IR &\textbf{2.2$\pm$2.3 }  &\textbf{2.2$\pm$1.6} &\textbf{2.2$\pm$2.0}  \\ \hline
\end{tabular}
}
\label{tab:ablation_modalities}
\end{minipage}
\begin{minipage}[t]{0.5\linewidth}
\makeatletter\def\@captype{table}\makeatother\caption{Comparison of fusion strategies.}
\scalebox{0.83}{
\begin{tabular}{|c|c|c|c|} \hline
    Method   & APCER(\%) & BPCER(\%) & ACER(\%) \\ \hline \hline
    NHF      & 25.3$\pm$12.2  & 4.4$\pm$3.1   & 14.8$\pm$6.8  \\ \hline
    PSMM-WoBF & 12.7$\pm$0.4  & 3.2$\pm$2.3   & 7.9$\pm$1.3  \\ \hline
    PSMM-Net &  \textbf{2.2$\pm$2.3}  & \textbf{2.2$\pm$1.6}   & \textbf{2.2$\pm$2.0}  \\ \hline
\end{tabular}
}
\label{tab:ablation_NHF_pSMM}
\end{minipage}
\end{table}

{\flushleft \textbf{Multiple Modalities.}} In order to show the effect of analysing a different number of modalities, we evaluate one modality (RGB), two modalities (RGB and Depth), and three modalities (RGB, Depth and IR) on PSMM-Net. As shown in Fig.~\ref{fig:static_dynamic}, the PSMM-Net contains three SD-Nets and one shared branch. When only RGB modality is considered, we just use one SD-Net for evaluation. When two or three modalities are considered, we use two or three SD-Nets and one shared branch to train the PSMM-Net model, respectively. Results are shown in Table~\ref{tab:ablation_modalities}. The best results are obtained when using all three modalities: $2.2\%$ of APCER, $2.2\%$ of BPCER and $2.2\%$ of ACER. These results show that multi-modal information has a significant effect in alleviating ethnic bias, mainly because of the smaller differences in skin color of different ethnicities in the IR modality.

\begin{figure}[t]
\begin{center}
\includegraphics[width=0.7\linewidth]{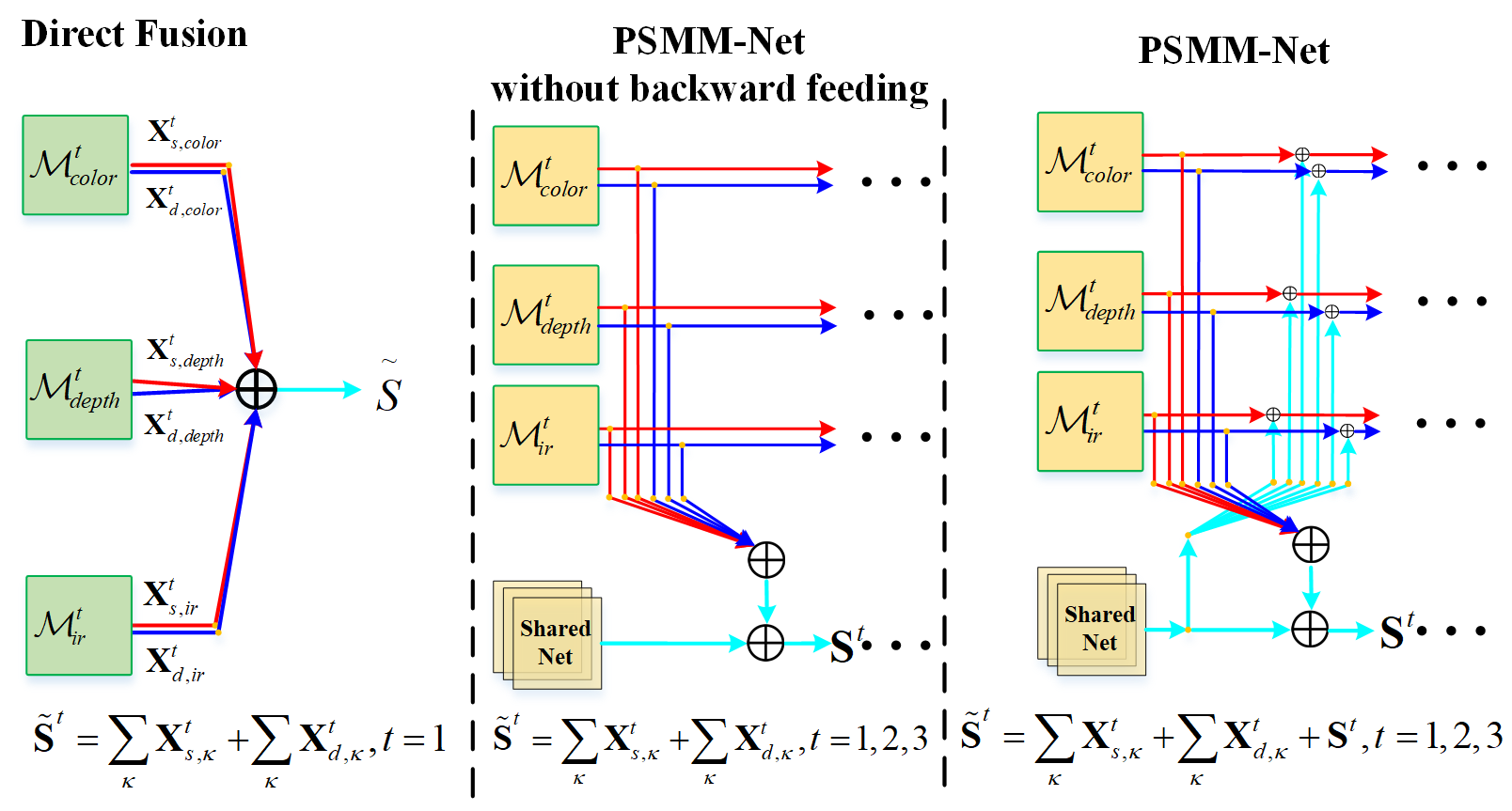}
\end{center}
\caption{Comparison of network units for multi-modal fusion strategies. From left to right: NHF, PSMM-NET-WoBF and PSMM-Net. The fusion process for the $t^{th}$ feature level of each strategy is shown at the bottom.}
\label{fig:modality_fusion_ways}
\end{figure}

{\flushleft \textbf{Fusion Strategy.}} In order to evaluate the performance of PSMM-Net, we compare it with other two variants:  Naive halfway fusion (NHF) and PSMM-Net without backward feeding mechanism (PSMM-Net-WoBF). As shown in Fig.~\ref{fig:modality_fusion_ways}, NHF combines the modules of different modalities at a later stage (\ie, after ${\cal M}^{1}_{\kappa}$ module) and PSMM-Net-WoBF strategy removes the backward feeding from PSMM-Net. The fusion comparison results are shown in Table~\ref{tab:ablation_NHF_pSMM}, showing higher performance of the proposed PSMM-Net with information exchange and interaction mechanism among SD-Nets and the shared branch.

\subsection{Methods Comparison}
{\flushleft \textbf{CASIA-SURF.}} The comparison results are show in Table~\ref{tab:casiasurf_result}. The performance of the PSMM-Net is superior to the ones of the competing multi-modal fusion methods, including Halfway fusion~\cite{DBLP:conf/cvpr/abs-1812-00408}, single-scale SE fusion~\cite{DBLP:conf/cvpr/abs-1812-00408}, and multi-scale SE fusion~\cite{zhang2019casiasurf}. When compared with~\cite{DBLP:conf/cvpr/abs-1812-00408,zhang2019casiasurf}, PSMM-Net improves the performance by at least $0.4\%$ for ACER. When the PSMM-Net is pretrained on CeFA, it further improves performance. Concretely, the performance of $TPR@FPR=10^{-4}$ is increased by $2.4\%$ when pretraining with the proposed CeFA dataset. The comparison results not only illustrate the superiority of our algorithm for multi-modal data fusion, but also show that our CeFA alleviates the bias of attack pattern to a certain extent.

\begin{table}[]
\begin{minipage}[t]{1.0\linewidth}
\centering
\makeatletter\def\@captype{table}\makeatother\caption{Comparison of the proposed method with three fusion strategies. All models are trained and tested on the CASIA-SURF. '()' means the method is trained from a specific dataset: S(CASIA-SURF), D(Data), C(CeFA). Best results are bolded.}
\centering
\scalebox{0.85}{
\begin{tabular}{|c|c|c|c|c|c|c|}
\hline
\multirow{2}{*}{Method} & \multicolumn{3}{c|}{TPR (\%)} & \multirow{2}{*}{APCER (\%)} & \multirow{2}{*}{BPCER (\%)} & \multirow{2}{*}{ACER (\%)}\\ \cline{2-4}
    	& @FPR=$10^{-2}$ &@FPR=$10^{-3}$ &@FPR=$10^{-4}$ & &  & \\ \hline \hline
    NHF~\cite{DBLP:conf/cvpr/abs-1812-00408} &89.1 &33.6 &17.8 &5.6 &3.8 &4.7 \\ \hline
    Single-scale SEF~\cite{DBLP:conf/cvpr/abs-1812-00408} &96.7 &81.8 &56.8 &3.8 &1.0 &2.4\\ \hline
    Multi-scale SEF~\cite{zhang2019casiasurf} &99.8 &98.4 &95.2 &1.6 &0.08 &0.8 \\	\hline
    PSMM-Net       &\textbf{99.9} &99.3 &96.2 &0.7 &0.06 &0.4  \\	\hline
    PSMM-Net(C) &\textbf{99.9} &\textbf{99.7} &\textbf{97.6} &\textbf{0.5} &\textbf{0.02} &\textbf{0.2}  \\	\hline
\end{tabular}
}
\label{tab:casiasurf_result}
\end{minipage}
\\
\begin{minipage}[t]{0.5\linewidth}
\centering
\makeatletter\def\@captype{table}\makeatother\caption{Comparisons on SiW. 'P' and 'Pr.' denote protocol and pretrain, respectively.}
\scalebox{0.72}{
\begin{tabular}{|c|c|c|c|c|c|}
\hline
    P & Method & APCER (\%) & BPCER (\%) & ACER (\%) & Pr. \\
    \hline
    \hline
    \multirow{6}{*}{1} & BAS~\cite{Liu2018Learning} & 3.58 & 3.58 & 3.58 & \multirow{3}{*}{No}\\
    \cline{2-5} & TD-SF~\cite{FASTD2018arxiv}       & 1.27 & \textbf{0.83} & 1.05  & \\
    \cline{2-5} & STASN~\cite{yang2019face}         & - & - & 1.00  & \\
    \cline{2-5} & SD-Net                            & \textbf{0.14} & 1.34 & \textbf{0.74} &  \\
    \cline{2-6} & \begin{tabular}[c]{@{}c@{}}TD-SF(S)\end{tabular}
    & 1.27 & \textbf{0.33} & 0.80  & \multirow{4}{*}{Yes}\\
    \cline{2-5} & STASN(D)        & - & - & \textbf{0.30}  &  \\
    \cline{2-5} & SD-Net(C)       & \textbf{0.21} & 0.50 & 0.35 &  \\
    \hline
    \hline

    \multirow{6}{*}{2} & BAS & 0.57$\pm$0.69 & 0.57$\pm$0.69 & 0.57$\pm$0.69 & \multirow{3}{*}{No}\\
    \cline{2-5} & TD-SF       & 0.33$\pm$0.27  & \textbf{0.29$\pm$0.39} & 0.31$\pm$0.28  & \\
    \cline{2-5} & STASN  & - & -  & 0.28$\pm$0.05 & \\
    \cline{2-5} & SD-Net & \textbf{0.25$\pm$0.32} & \textbf{0.29$\pm$0.34} & \textbf{0.27$\pm$0.28} & \\
    \cline{2-6} & \begin{tabular}[c]{@{}c@{}}TD-SF(S) \end{tabular}
    & \textbf{0.08$\pm$0.17}  & 0.25$\pm$0.22 & 0.17$\pm$0.16  & \multirow{4}{*}{Yes}\\
    \cline{2-5} & STASN(D)    & - & - & \textbf{0.15$\pm$0.05} & \\
    \cline{2-5} & SD-Net(C)   & 0.09$\pm$0.17 & \textbf{0.21$\pm$0.25} & \textbf{0.15$\pm$0.11} & \\
    \hline
    \hline
    \multirow{6}{*}{3} & BAS & 8.31$\pm$3.81  & 8.31$\pm$3.81  & 8.31$\pm$3.81 & \multirow{3}{*}{No}\\
    \cline{2-5} & TD-SF      & 7.70$\pm$3.88  & \textbf{7.76$\pm$4.09}   & 7.73$\pm$3.99 & \\
    \cline{2-5} & STASN      & - & -  & 12.10$\pm$1.50 & \\
    \cline{2-5} & SD-Net     & \textbf{3.74$\pm$2.15} & 7.85$\pm$1.42 & \textbf{5.80$\pm$0.36} & \\
    \cline{2-6} & \begin{tabular}[c]{@{}c@{}}TD-SF(S)\end{tabular}
    & 6.27$\pm$4.36  & \textbf{6.43$\pm$4.42} & 6.35$\pm$4.39  & \multirow{4}{*}{Yes}\\
    \cline{2-5} & STASN(D)      & - & -  & 5.85$\pm$0.85 & \\
    \cline{2-5} & SD-Net(C)           & \textbf{2.70$\pm$1.56} & 7.10$\pm$1.56 & \textbf{4.90$\pm$0.00} & \\
    \hline
\end{tabular}
}
\label{tab:siw_result}
\end{minipage}
\begin{minipage}[t]{0.5\linewidth}
\centering
\makeatletter\def\@captype{table}\makeatother\caption{Comparisons on OULU-NPU. 'P' and 'Pr.' denote protocol and pretrain, respectively.}
    \scalebox{0.72}{
    \begin{tabular}{|c|c|c|c|c|c|}
    \hline
    P & Method & APCER (\%) & BPCER (\%) & ACER (\%) & Pr.\\
    \hline
    \hline
    \multirow{6}{*}{1} & BAS~\cite{Liu2018Learning} & 1.6 & \textbf{1.6} & 1.6 & \multirow{4}{*}{No}\\
    \cline{2-5} & Ds~\cite{Jourabloo2018Face}     & \textbf{1.2} & 1.7 & \textbf{1.5} & \\
    \cline{2-5} & STASN~\cite{yang2019face}       & \textbf{1.2} & 2.5 & 1.9 & \\
    \cline{2-5} & SD-Net                          & 1.7 & 1.7 & 1.7 & \\

    \cline{2-6} & STASN(D)     & 1.2 & \textbf{0.8} & \textbf{1.0} & \multirow{2}{*}{Yes} \\
    \cline{2-5} & SD-Net(C)    & \textbf{1.0} & 1.7 & 1.4 & \\
    \hline
    \hline
    \multirow{6}{*}{2} & BAS & \textbf{2.7} & 2.7 & 2.7 & \multirow{4}{*}{No}\\
    \cline{2-5} & STASN        & 4.2 & \textbf{0.3} & \textbf{2.2} &\\
    \cline{2-5} & SD-Net       & 2.8 & 2.2 & 2.5 &\\
    \cline{2-6} & STASN(D)     & \textbf{1.4} & \textbf{0.8} & \textbf{1.1} & \multirow{2}{*}{Yes}\\
    \cline{2-5} & SD-Net(C)    & \textbf{1.4} & 2.5 & 1.9 &\\
    \hline
    \hline

    \multirow{6}{*}{3} & BAS & \textbf{2.7$\pm$1.3} & 3.1$\pm$1.7 & 2.9$\pm$1.5 & \multirow{4}{*}{No}\\
    \cline{2-5} & STASN       & 4.7$\pm$3.9 & 0.9$\pm$1.2 & 2.8$\pm$1.6 & \\
    \cline{2-5} & SD-Net      & \textbf{2.7$\pm$2.5} & \textbf{1.4$\pm$2.0} & \textbf{2.1$\pm$1.4} & \\
    \cline{2-6} & STASN(D)    & \textbf{1.4$\pm$1.4} & 3.6$\pm$4.6 & 2.5$\pm$2.2 & \multirow{2}{*}{Yes}\\
    \cline{2-5} & SD-Net(C)   & 2.7$\pm$2.5 & \textbf{0.9$\pm$0.9} & \textbf{1.8$\pm$1.4} & \\
    \hline
    \hline
    \multirow{6}{*}{4} & BAS & 9.3$\pm$5.6  & 10.4$\pm$6.0  & 9.5$\pm$6.0 & \multirow{4}{*}{No}\\
    \cline{2-5} & STASN       & 6.7$\pm$10.6 & 8.3$\pm$8.4   & 7.5$\pm$4.7 & \\
    \cline{2-5} & SD-Net      & \textbf{4.6$\pm$5.1}  & \textbf{6.3$\pm$6.3}   & \textbf{5.4$\pm$2.8} & \\
    \cline{2-6} & STASN(D) & \textbf{0.9$\pm$1.8}  & \textbf{4.2$\pm$5.3}   & \textbf{2.6$\pm$2.8} & \multirow{2}{*}{Yes}\\
    \cline{2-5} & SD-Net(C)   & 5.0$\pm$4.7  & 4.6$\pm$4.6   & 4.8$\pm$2.7 & \\
    \hline
\end{tabular}
}
\label{tab:oulu_result}
\end{minipage}
\end{table}

{\flushleft \textbf{SiW and OULU-NPU.}} Results for these two dataset are shown in Table~\ref{tab:siw_result} and~\ref{tab:oulu_result}, respectively. We compare the proposed SD-Net with other methods without pretraining. Our method achieves the best results (a lower ACER value indicates better performance) on all protocols of the SiW and protocol $3$ and $4$ of the OULU-NPU. The experimental results show that our SD-Net combined with the dynamic image generated by the rank pooling algorithm can effectively capture features related to motion difference between the real face and the fake one.

Last but not least, using the proposed dataset to pre-train our baseline method significantly improves its ACER performance in most of protocols. In Protocol $2$ and $3$ of SiW, our method trained on the CeFA dataset performs the best among all models. Note that the STASN (Data)~\cite{yang2019face} used a large private dataset to pretrain. Similar conclusions can be drawn from the OULU-NPU experiment. These results demonstrate the effectiveness and generalization capability of the CeFA dataset, and suggest SOTA methods can be further improved by using our CeFA dataset for pre-training.

\subsection{Visualization and Analysis}
\begin{figure}[t]
\begin{center}
\includegraphics[width=1.0\linewidth]{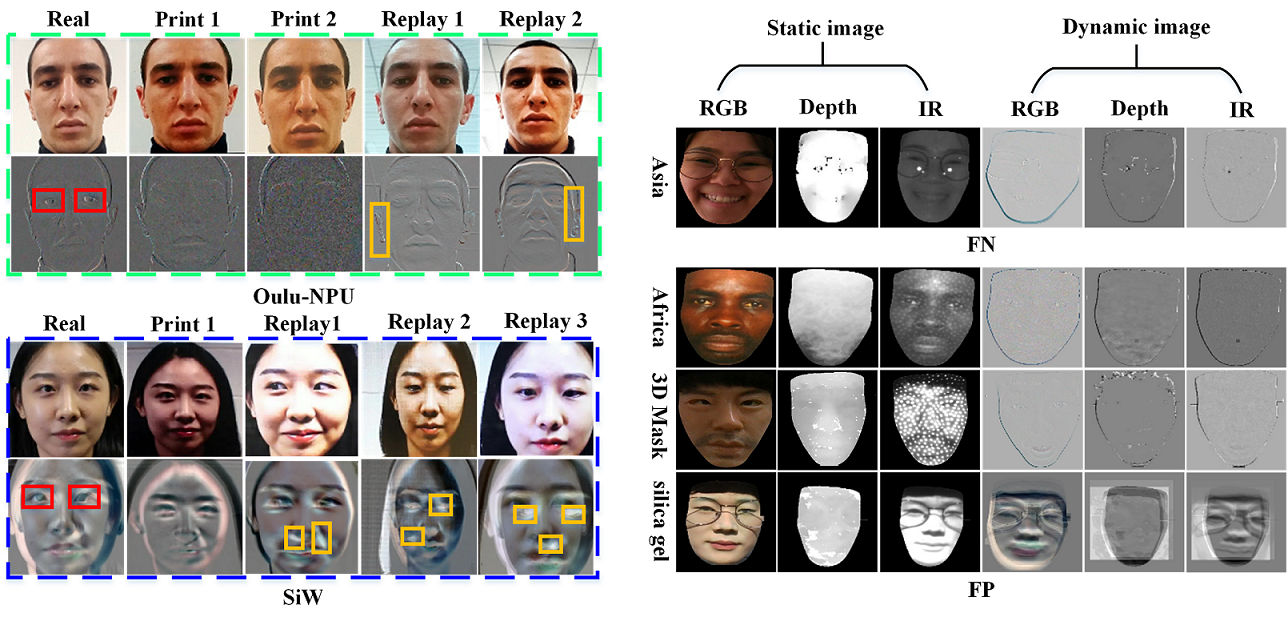}
\caption{(a) RGB samples (the first and third row) with their corresponding dynamic image (the second and fourth row), and their labels in the top of each column. (b) Misclassified examples. First three columns are three modal static images, and last three columns correspond dynamic image. The first row are Central Asia real faces and the last three rows are attack samples: print attack (Africa), 3D mask, and silicone mask. FP: False Positive; FN: False Negative.}
\label{fig:race_oulu_siw_rp}
\end{center}
\end{figure}

{\flushleft \textbf{Dynamic images.}}
Given a video, we map a sequence of $7$ frames into a dynamic image by using rank pooling. Some samples are shown in Fig.~\ref{fig:race_oulu_siw_rp}(a). As for SiW and OULU-NPU datasets, the eye part (red box) of the real sample is more realistic than print or replay attack, while more speckles (orange box) caused by specular reflections are included in the replay attack. Our SD-Net can capture these discriminative dynamic features.
{\flushleft \textbf{Misclassified Samples.}}
Some misclassified samples of our baseline on CeFA are shown in Fig.~\ref{fig:race_oulu_siw_rp}(b). Visually from static image, it is very difficult to distinguish the type of the sample from RGB and IR modalities. Furthermore, the depth modality of a 3D attack shows to be extremely similar to the real face. 

\section{Conclusion}
In this paper, we release the largest face anti-spoofing dataset up to date in terms of modalities, number of subjects and attack types. More importantly, CeFA is the only public face anti-spoofing dataset with ethnic labels. Specially, we define four protocols to study the generalization performance of face anti-spoofing algorithms. Based on the proposed dataset, we provide a baseline by designed a partially shared PSMM-Net to learn complementary information from multi-modal data in videos, in which a SD-Net aims to learn both static and dynamic features from single modality. Extensive experiments validate the utility of our algorithm and the challenges of the released CeFA dataset.

\section{Acknowledgement}
This work has been partially supported by the Chinese National Natural Science Foundation Projects $\#$61961160704, $\#$61876179, $\#$61872367, Science and Technology Development Fund of Macau (Grant No. 0025/2018/A1), the Spanish project TIN2016-74946-P (MINECO/FEDER, UE) and CERCA Programme / Generalitat de Catalunya, and by ICREA under the ICREA Academia programme. We acknowledge Surfing Technology Beijing co., Ltd (www.surfing.ai) to provide us this high quality dataset. We also acknowledge the support of NVIDIA Corporation with the donation of the GPU used for this research.


\bibliographystyle{splncs03}
\bibliography{egbib}
\end{document}